\documentclass{article}

\usepackage{arxiv}

\usepackage[utf8]{inputenc} 
\usepackage[T1]{fontenc}    
\usepackage{hyperref}       
\usepackage{url}            
\usepackage{booktabs}       
\usepackage{amsfonts}       
\usepackage{nicefrac}       
\usepackage{microtype}      
\usepackage{lipsum}		
\usepackage{graphicx}
\usepackage{algpseudocode}
\usepackage{algorithm}
\usepackage[cmex10]{amsmath}
\usepackage{algpseudocode}
\usepackage{mdwmath}
\usepackage{mdwtab}
\usepackage{algorithm}
\usepackage{nccmath}
\usepackage{amsmath}

\makeatletter
\def\BState{\State\hskip-\ALG@thistlm}
\makeatother

\title{Follow Pedro! An Infrared-based Person-Follower using Nonlinear Optimization}

\author{
  Pedro A. Pe\~na\thanks{https://www.pedroapena.com} \\
  Robotics\\
  Toyota Research Institute\\
  Cambridge, MA. 02139 \\
  \texttt{pedro.pena.ctr@tri.global} \\
   \And
 Toffee Albina \\
  Robotics\\
  Toyota Research Institute\\
  Cambridge, MA. 02139\\
  \texttt{toffee.albina@tri.global} \\
}

\begin{document}
\maketitle

\begin{abstract}
We used ROS2 as a platform to conduct AI research for developing a Follow-Me capability as a proof-of-concept on a wheeled robot, demonstrating that AI research is possible in the ROS2 framework.
We developed a complete system that uses perception and navigation components based on a sensor suite of fisheye cameras, lidar, and IMU running on an ARM64 Embedded Linux platform that runs ROS2 natively. The perception package detects AR markers and/or IR beacons to track a person. The tracker uses AI algorithms such as particle filters and nonlinear optimization to extract the SE(3) information of the 2D feature.
\end{abstract}


\section{Introduction}
We propose a real-time integrated system that utilizes minimal amount of computational power that can enable robots to follow a person through an AR marker or IR beacon. Although a plethora of research has been done on this topic, they either rely on machine learning \cite{chen2017socially,pang2019efficient} or human behavior models \cite{ferrer2013robot,bera2017sociosense,randhavane2019pedestrian,garrell2012cooperative} and computationally expensive, i.e., high spatial and temporal complexity, algorithms \cite{kim2018architecture,priyandoko2018human,tsai2018real,tee2018improved} in which data and high-end hardware for computational power is required. This paper is meant for current ROS1 research teams, researchers looking for a platform for AI/Robotics research, and researchers working on embedded systems that can benefit from inexpensive hardware to create an integrated system that is able to follow a person. Our system utilizes an inexpensive ARM64 processor that can run Linux and ROS2 natively and is able to detect and follow a person reliably. In section \ref{sec:follower}, we will discuss the algorithm used to follow a person given a detection. Then in section \ref{sec:marker}, we will explain how a marker is detected using an AR marker, and in section \ref{sec:beacon} we explain the IR beacon detection algorithm that utilizes a nonlinear optimization to infer the 3D pose from a 2D triangle on an image. We will then show how our method was used on a Turtlebot platform to follow a person and a robot in section \ref{sec:application}.

\section{Follower}
\label{sec:follower}
The Follow-Me algorithm consists of detectors and a follower component. For this research, we developed two detection algorithms:
\begin{itemize}
\item AR Marker
\item IR Beacon with active 3 infrared transmitters in a triangle configuration on
black
\end{itemize}

Both of these detectors output a range and bearing in the robot's base frame enabling the Follow-Me implementation to be modular. In this way we can develop detectors and followers separately from each other.
The follower uses a PID-controller to get within a specified radius of a person. Consequently, the velocity of the robot is proportional to the distance from the person to the robot. Therefore as the robot gets close to the person, it will slow down until it has reached its equilibrium. In order to reach its target, it uses relative odometry where it does not depend on a global frame of reference. When we receive a range and bearing from a detector, i.e., a detection of the human the robot is following, we verify that the range and bearing (i.e. defined as vector $\mathbf{\xi}_{rb}$) of the person the robot is following is greater than some value  $\mathbf{\epsilon}_{rb}$. If this is the case, the robot will move towards the position of the human. As stated before, we use a PID controller where the error is the range and bearing of the person. Since we want to avoid the robot bumping into the human it is following, we subtract a predetermined distance vector, $\mathbf{d}$, from the range so that the robot is always at a distance $\mathbf{d}$ from the person. When the range and bearing are below some threshold $\mathbf{\epsilon}_{rb}$, the robot will stop moving. The algorithm is as follows:

  \begin{equation}
    Follow =
    \begin{cases}
      GoToHuman(\mathbf{\xi}_{rb} - \mathbf{d}), & \text{if}\ \lVert \mathbf{\xi}_{rb} - \mathbf{d} \rVert  \geq \mathbf{\epsilon}_{rb} \\
      Stop(), & \text{otherwise}
    \end{cases}
  \end{equation}
%


\section{AR-Marker Detector}
\label{sec:marker}
\begin{figure}
  \centering
\includegraphics[width=0.7\textwidth]{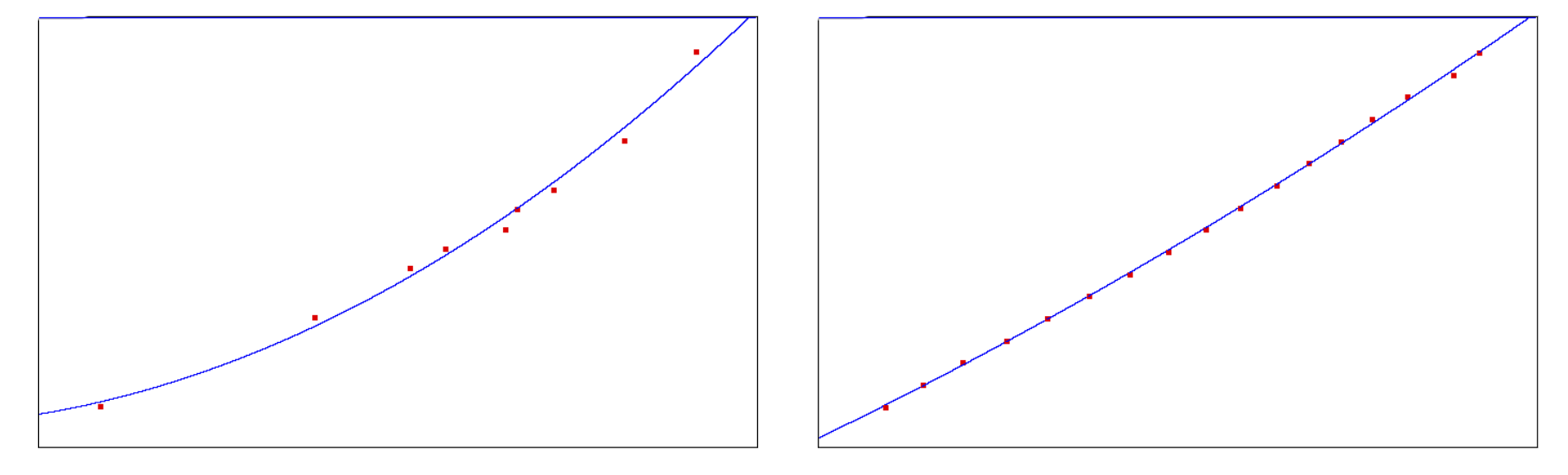}
\includegraphics[width=0.35\textwidth]{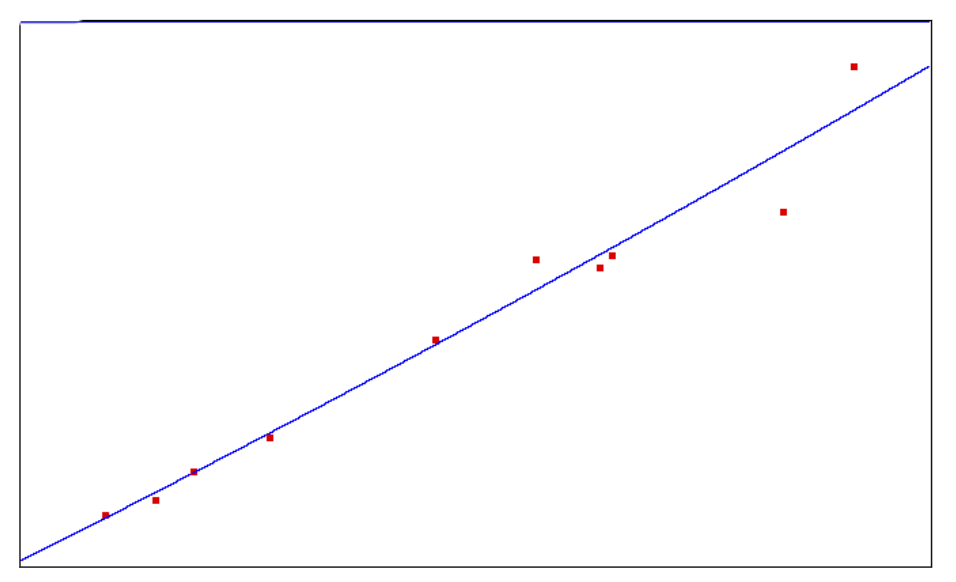}
  \caption{The graphs show the comparison of fisheye space vs Euclidean space. The top left graph shows bearing data fitting for the rear camera and the top right graph shows bearing data fitting for the front camera. The bottom graph shows the range data fitting for the front camera.  Note: the range data fitting was only done for the front camera because the robot does not move towards the robot from the rear (i.e. the robot will rotate towards the person). The range of the data is \textasciitilde$(-\pi, \pi)$}
  \label{fig:fitdata}
\end{figure}

The marker detector uses the ArUco library \footnote{\url{https://docs.opencv.org/trunk/d5/dae/tutorial_aruco_detection.html}} to find the 3D pose of an AR marker tag. To utilize fisheye cameras, for the advantage of the field of view, the 3D pose given by ArUco may not be correct because of distortions of the camera in the border of the lens. Based on empirical observations, there is a mapping between fisheye space and Euclidean space.
The camera is sampled with the marker in different configurations and a transformation from fisheye position to a Euclidean position is fitted with a quadratic equation; the graphs in figure \ref{fig:fitdata} show the relationship between these two spaces for two different cameras.The sampling of configurations can be done by recording the pose of the marker and the output of ArUco; this needs to be done for every camera. After we learn this map, the range and bearing is given in the robot's base frame. 

%

%
%
\begin{figure}
  \centering
  \href{https://www.youtube.com/watch?v=qOWwjKRiL9o}{\includegraphics[width=0.4\textwidth]{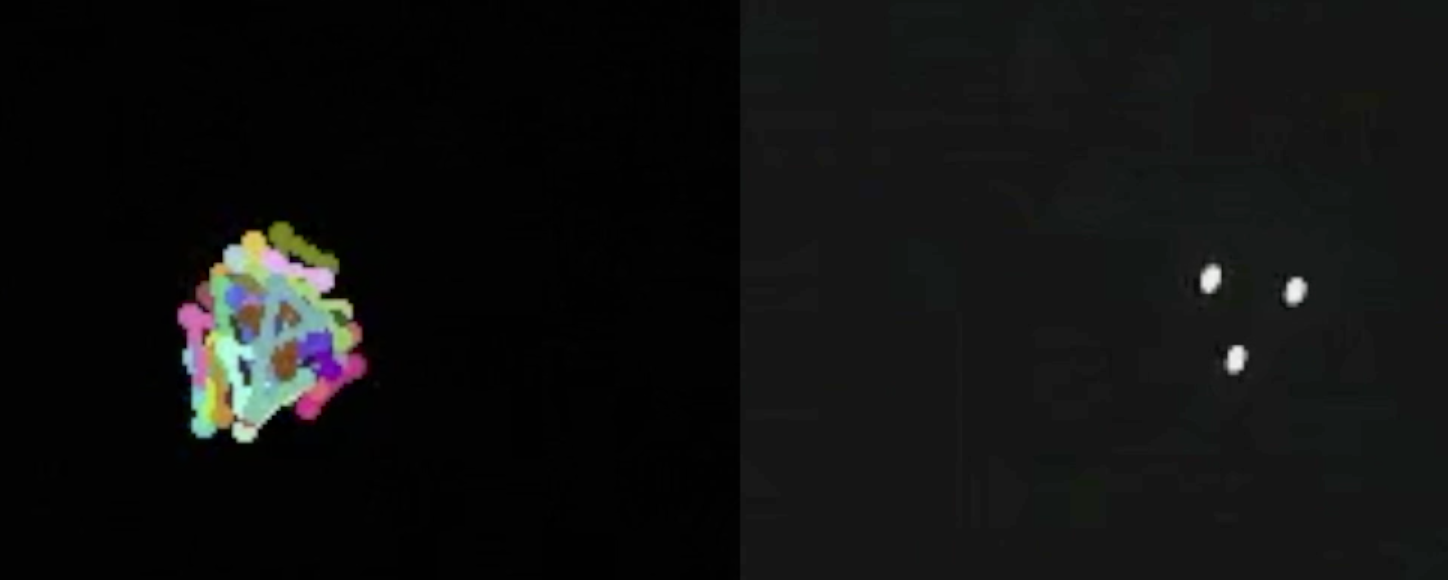}}
  \caption{ The right image shows the perception of a 3D printed black puck containing three IR transmitters in a triangle formation from a fisheye camera fitted with an IR filter. The left image shows a particle filter tracking the IR beacon.}
  \label{fig:ir}
\end{figure}
\section{IR-Beacon Detector}
\label{sec:beacon}
For this detector, an IR filter is fitted to the lenses of the fisheye cameras to solely perceive the IR light. A 3D printed black puck containing three IR transmitters in a triangle formation is attached to the leader; figure \ref{fig:ir} shows a perception of the IR beacon by a fisheye camera. The input of the algorithm is the camera image, and we need to perceive three bright blobs that form a triangle on a black background. The algorithm proceeds as follows:
\begin{itemize}
\item Threshold and get only the pixel values above the threshold. This threshold function will give us the brightest pixels.
\item Get the contours in the image and approximate the contours with polynomials.
\item Find circles that bound these polynomials - these are the potential IR beacons.
\item We then do a permutation of tuples of three circles to form a triangle. The
triangles that satisfy the following constraints become the candidate triangles for the multivariate Newton's method:
\begin{itemize}
\item The difference of the side between the first and second beacon and the side between the first and third beacon need to be less than a side threshold
\item The radius of the first, second, and third beacon needs to be below a threshold
\item The lengths of the sides needs to be below a threshold
\item The pixel value midway between the triangle sides needs to be above a pixel threshold
\item The pixel value in the center of the triangle needs to be above a threshold
\end{itemize}
\end{itemize}

The triangles that pass these constraints can be used for the multivariate Newton's method. The multivariate Newton's method is used to solve the roots of nonlinear equations, and in our case, we are using it to find the P3P correspondence of the triangle in image space to Euclidean space with the equation: 
\begin{equation}
f(\mathbf{x}_n) =\mathbf{u} = \mathbf{F} \cdot \mathbf{T} \cdot \mathbf{x},
\label{eq:equation}
\end{equation}
where $\mathbf{u}$ is the image coordinates in pixel space, $\mathbf{F}$ is the camera matrix, $\mathbf{T}$ is the transformation with respect to the optical frame (this is what we are looking for), and $\mathbf{x}$ is the beacon pose in 3D space relative to a coordinate frame in Euclidean space (we chose the center point of the triangle formed by the IR transmitters). The equation is formulated
as a difference equation where the steps are discrete:
\begin{equation}
F(x) = x_n - \frac{f(x_n)}{f'(x_n)}, \quad x(0) = c,
\end{equation}
where c is the initial condition. This equation iterates until the following constraint is satisfied: $F(x_n) - F(x_{n-1}) \leq \epsilon$, i.e., this means we have hit a fixed point which directly translates to the roots of equation \ref{eq:equation}. It is important to note that the state of $\mathbf{T}$ is $SE(3)$ which makes the difference equation multivariate. Hence, the equation is: 
\begin{equation}
\mathbf{x}_{n+1} = \mathbf{x}_n  - \mathbf{J^{+}} \cdot f(\mathbf{x}_n),
\end{equation}
where $\mathbf{J^{+}}$ represents the pseudoinverse of the Jacobian matrix. In order to smooth the output of the system and compensate for any perception or numerical errors, the detector injects uncertainty into the system with a particle filter to keep track of the IR beacon in Euclidean space. The motion model used for the prediction step is: 
\begin{equation}
\mathbf{x}_{n+1} = \mathbf{x}_{n} + \mathbf{u}(t) + \mathbf{v}(t),
\end{equation}
where $\mathbf{u}(t)$ represents the control input of the robot, $\mathbf{x}_{n}$ is particle $n$'s pose in 3D space and $\mathbf{v}(t)$ represents the velocity of the IR beacon in world space. The mean particle pose is used as the output for the range and bearing. Figure \ref{fig:ir} shows the IR beacon tracked by the particle filter.

A video of an IR beacon in a triangle configuration running the multivariate Newton's method can be found here:
\begin{center}
  \url{https://www.youtube.com/watch?v=qOWwjKRiL9o}
\end{center}
\begin{figure}[h]
  \centering
\includegraphics[width=0.4\textwidth]{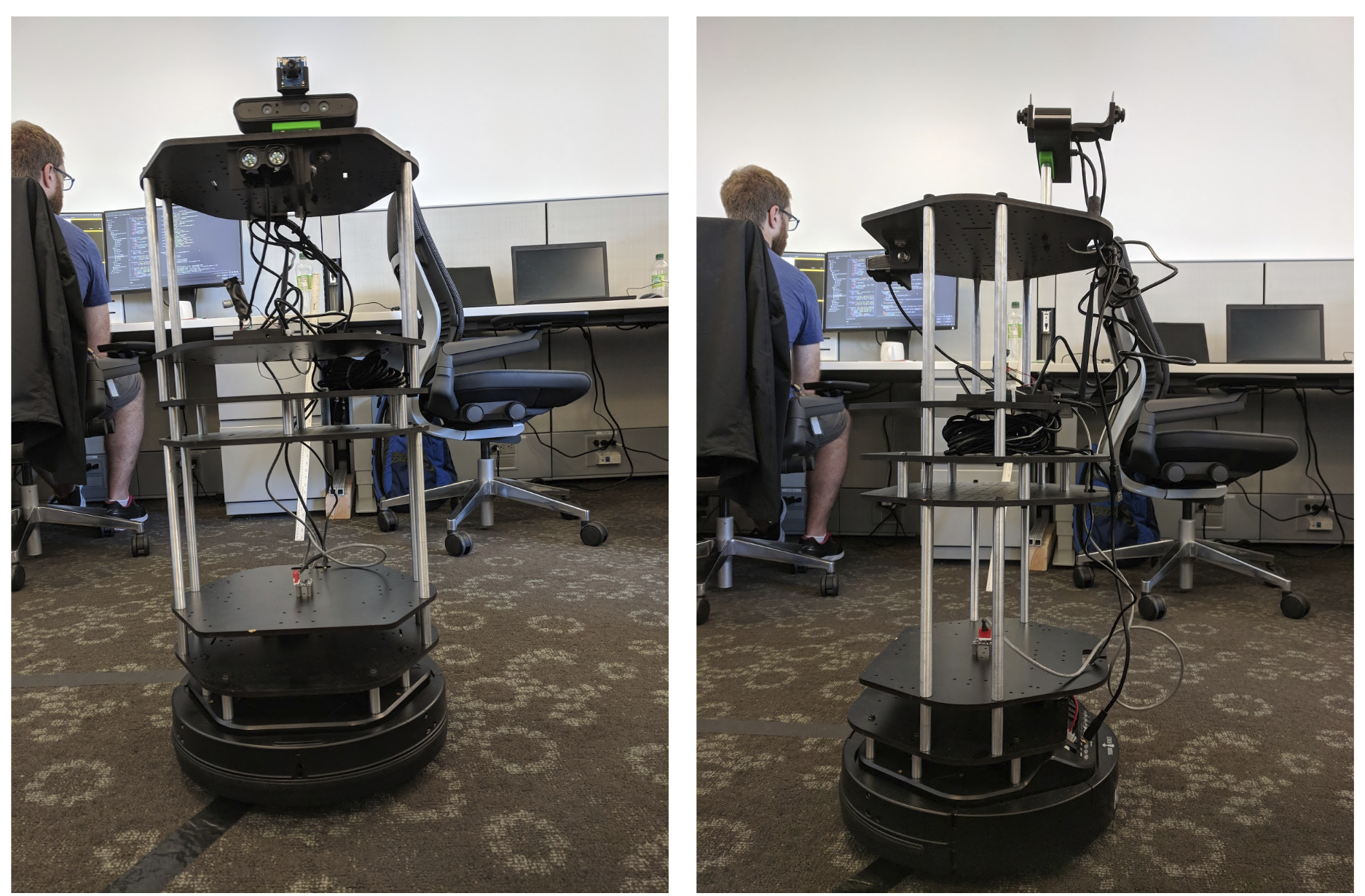}
  \caption{TurtlebotRex used for Follow-Me and built with a Kobuki base and two sets of Turtlebot 2 hardware stacked vertically to better situate the cameras for detection. The top of the robot's Orbbec Astra camera has two fisheye cameras, e.g., one facing front and another camera facing rear. The Bosch BNO055 IMU is on the bottom of the robot. There is also a Garmin Lidar Lite V3HP by the bottom of the Orbbec Astra camera. All the sensors and algorithms are powered by the ROCK64, an ARM64 processor. }
  \label{fig:tb}
\end{figure}
\section{Application}
\label{sec:application}
\begin{figure}[h]
  \centering
\includegraphics[width=0.6\textwidth]{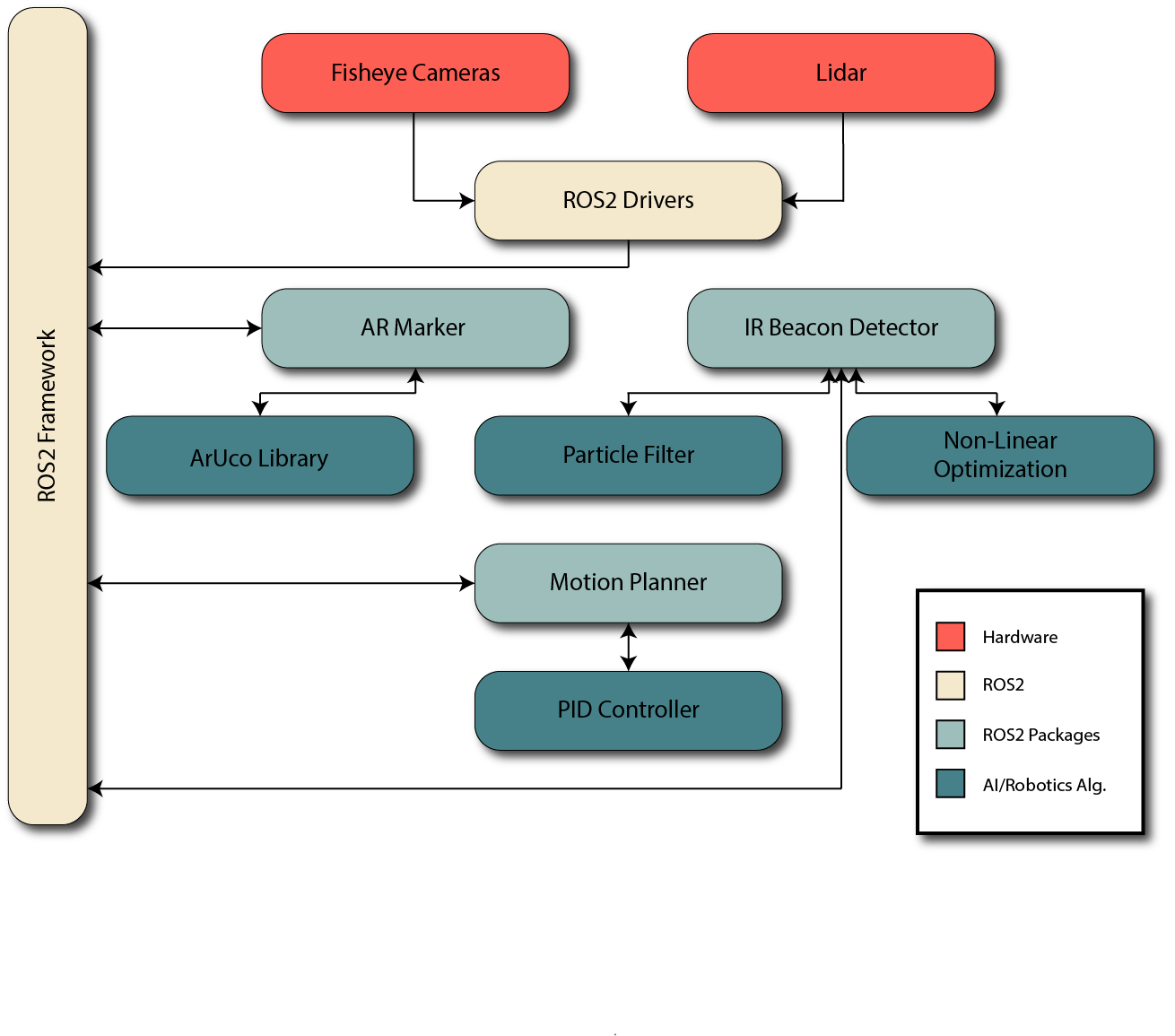}
\vspace{-14mm}
  \caption{A system diagram of the Follow-Me application.}
  \label{fig:sys}
\end{figure}

To evaluate the algorithm proposed in this paper, the follower and detectors were implemented on ROS2 using the Turtlebot 2 hardware and software stack \footnote{\url{https://github.com/ros2/turtlebot2_demo}} so that we can communicate with the Kobuki base for motor control. Because the height of Turtlebot 2 is short and the application requires detections of humans, we built a robot that is composed of two Turtlebot 2 platforms stacked on top of each other to situate the cameras at a more proper height: \emph{TurtlebotRex}. For the camera system setup, the original Turtlebot 2 came with an Orbbec Astra Camera for depth and RGB data. The Orbbec Astra Camera contained the following drawbacks:
\begin{itemize}
\item Low FPS depth stream, and
\item Limited FOV(60 degrees horizontal).
\end{itemize}
To eliminate these drawbacks, two fisheye cameras are used, one for the front view and back view, creating an almost full view, omnidirectional,  around the robot. The limitations of a fisheye camera such as distortion are considered in the detectors as discussed in sections \ref{sec:marker} and \ref{sec:beacon}. The depth sensor used for TurtlebotRex is a Garmin Lidar Lite V3HP, and an IMU, Bosch BNO055, is also used to filter odometry. The processing unit is an ARM64 processor, ROCK64, which has 4 GB of memory  and runs ROS2 Bouncy Bolson natively. Figure \ref{fig:tb} shows how all the components are arranged on TurtlebotRex and a system diagram is shown in figure \ref{fig:sys}.

A video of TurtlebotRex following a person with an AR marker can be found here:
\begin{center}
  \url{https://www.youtube.com/watch?v=COBOlq15_DU}
\end{center}

A video of  TurtlebotRex following a robot with an AR marker can be found here:
\begin{center}
  \url{https://www.youtube.com/watch?v=KU3pCqkwUrE}
\end{center}

A video of TurtlebotRex following a person with an IR beacon can be found here:
\begin{center}
  \url{https://www.youtube.com/watch?v=1ti0Bj0yDfI}
\end{center}
\section{Conclusion}
In this paper, we show a Follow-Me application using ROS2 where a physical robot, TurtlebotRex, was used. The Follow-Me application consists of modular components such as detectors, for perceptions of the follower, and a follower algorithm that enables a robot to follow its target. One of the detectors, IR beacon detector, uses a nonlinear optimization to correspond 2D features on an image of three IR beacons in a triangle configuration to their 3D position in the world. This application of robotics on ROS2 demonstrates that lightweight algorithms on inexpensive processors is possible with the correct technologies enabling interesting interactions with robots and humans.

\bibliographystyle{unsrt}  

\bibliography{follow}

\end{document}